\documentclass[lettersize,journal]{IEEEtran}
\usepackage{cite}
\usepackage{amsmath,amssymb,amsfonts,bm}
\usepackage{algorithmic}
\usepackage{graphicx}
\usepackage{textcomp}
\usepackage{xcolor}
\usepackage{pifont}
\usepackage{times}
\usepackage{latexsym}
\usepackage[ruled, linesnumbered]{algorithm2e}
\usepackage{algorithm2e,setspace}
\usepackage{epstopdf}
\usepackage{subfigure}
\usepackage{multirow}
\usepackage{booktabs}
\usepackage{mathrsfs} 
\usepackage{array}
\usepackage{makecell}
\usepackage{diagbox} 

\begin{document}
\title{Learning Relation-Specific Representations for Few-shot Knowledge Graph Completion}
\author{Yuling Li, 
	Kui Yu,
	Yuhong Zhang, 
	and~Xindong Wu
	        \thanks{
		This work is supported by the National Key Research and Development Program of China
		(under grant 2020AAA0106100).
	}
\thanks{Yuling Li, Kui Yu, Yuhong Zhang and Xindong Wu are with the Key Laboratory of
	Knowledge Engineering with Big Data of Ministry of Education, Hefei
	University of Technology, Hefei 230601, China, and also with the School
	of Computer Science and Information Engineering, Hefei University of
	Technology, Hefei 230601, China (e-mail:lyl95@mail.hfut.edu.cn, yukui@hfut.edu.cn, zhangyh@hfut.edu.cn, and xwu@hfut.edu.cn).}}


\maketitle

\begin{abstract}
	Recent years have witnessed increasing interest in few-shot knowledge graph completion (FKGC), which aims to infer unseen query triples for a few-shot relation using a few reference triples about the relation. The primary focus of existing FKGC methods lies in learning relation representations that can  reflect the common information shared by the query and reference triples. 
	To this end, these methods learn entity-pair representations from the direct neighbors of head and tail entities, and then aggregate the representations of reference entity pairs.
	However, the entity-pair representations learned only from direct neighbors may have low expressiveness when the involved entities have sparse direct neighbors or share a common local neighborhood with other entities.
	Moreover, merely modeling the semantic information of head and tail entities is insufficient to accurately infer their relational information especially when they have multiple relations.
	To address these issues,
	we propose a Relation-Specific Context Learning (RSCL) framework, which exploits graph contexts of triples to learn global and local relation-specific representations for few-shot relations.
	Specifically, we first extract graph contexts for each triple, which can provide long-term entity-relation dependencies.
	To encode the extracted graph contexts, we then present a hierarchical attention network to capture contextualized information of triples and highlight valuable local neighborhood information of entities.
	Finally, we design a hybrid attention aggregator to evaluate the likelihood of the query triples at the global and local levels.
	Experimental results on two public datasets demonstrate that 
	RSCL outperforms state-of-the-art FKGC methods.
\end{abstract}


\section{Introduction}
\label{sec:introduction}
A knowledge graph (KG) is essentially a multi-relational network composed of a large number of facts.
Each fact is organized as a triple (\textit{h}, \textit{r}, \textit{t}): \textit{(head entity, relation, tail entity)}, e.g., (\texttt{China}, \texttt{CountryCapital}, \texttt{Beijing}), indicating that Beijing is the capital of China.
Today, large scale KGs like YAGO \cite{suchanek2007yago}, NELL \cite{carlson2010toward}, and Wikidata \cite{vrandevcic2014wikidata} have been widely applied in various downstream applications, such as question \& answering \cite{saxena2020improving}, semantic search \cite{xiong2017explicit}, and information extraction \cite{bordes2013translating}.

Although typical KGs are large in size, they are always incomplete since the limited texts used to construct KGs cannot cover all the facts.
To solve this problem, researchers have proposed the KG completion task,
which aims to infer new triples by exploring latent semantics of entities and relations in existing triples.
Over the past decade, knowledge graph embedding (KGE) has
proven to be a powerful technique for the KG completion task \cite{bordes2013translating,trouillon2016complex,zhang2019iteratively}.
These KGE-based methods typically represent each entity and each relation in KGs as a low-dimensional vector and measure the plausibility of triples based on their semantic representations.
Despite their success, existing KGE-based models always require sufficient training triples for both entities and relations.
However, 
real-world KGs contain a large proportion of relations having only a limited number of triples.
Relations with a few triples are called few-shot relations.
For instance, about 10\% of relations in Wikidata \cite{vrandevcic2014wikidata} have no more than 10 triples \cite{chen2019meta}.
In addition,
for many practical applications such as social media and recommendation systems, KGs are always dynamically updated over time, and just a few triples are temporarily available for newly added relations.
Consequently, the performance of KGE-based models drops dramatically when they perform the KG completion task for the few-shot or newly added relations.


To tackle this issue,
a series of few-shot knowledge graph completion (FKGC) methods have been proposed to predict the missing tail entity $t$ in a query triple $(h,r,?)$ given only $K$ reference triples about the few-shot relation $r$ \cite{xiong2018one,sheng2020adaptive,chen2019meta,jiang2021metap,zhang2021gaussian,wang2021reform}.
The key idea behind FKGC methods is to learn a vector representation for each few-shot relation, which reflects the
common and shared information within one relation and can be transferred from reference triples to query triples.
To this end, 
most FKGC models learn the entity-pair representations by encoding the direct (one-hop) neighbors of head and tail entities, and then 
aggregate the representations of reference entity pairs to obtain the representation of the corresponding few-shot relation.

\begin{figure}[!t]
\centering	
\subfigure{\includegraphics[width=0.95\linewidth, trim=0 0 0 0, clip]{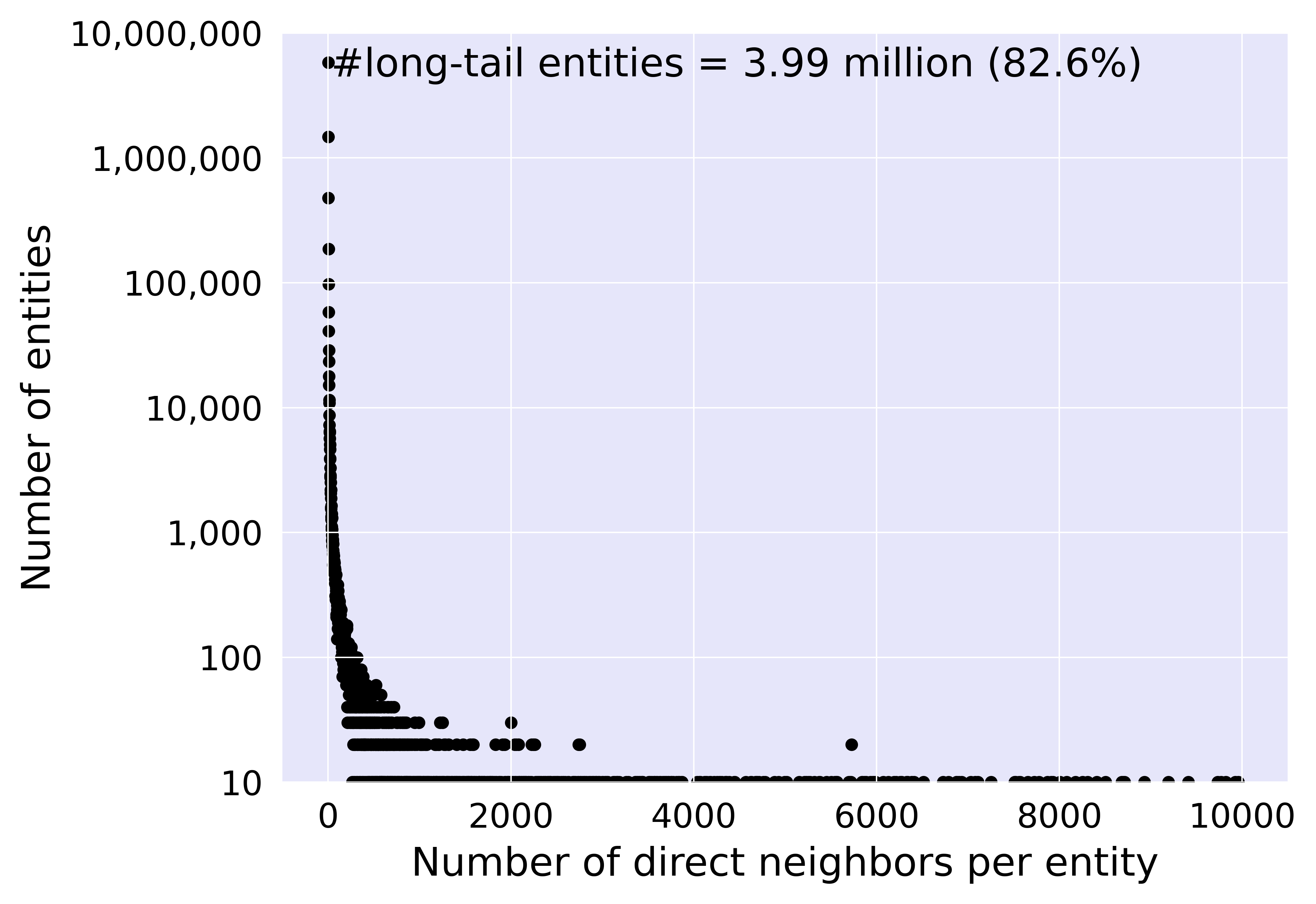}}
\caption{
	Distribution of number of entities versus number of direct neighbors per entity.
}
\label{fig1}
\end{figure}

However, these FKGC methods have two limitations:
(1) They learn representations of entity pair from a local view.
On the one hand, there are a large amount of entities having very few direct neighbors (a.k.a long-tailed entities) in real-world KGs. 
Figure \ref{fig1} shows that around 82.6\% of entities in Wikidata have only one direct neighbor.
These long-tailed entities receive limited neighbor information from their local neighborhoods, thus the representations of the involved entity pairs have low expressiveness.
On the other hand, 
there are many different entities sharing common direct neighbors in KGs \cite{guo2019learning}.
Therefore, exploiting direct neighbors is not sufficient to reveal the subtle differences between these entities, resulting in low-discrimination representations for the involved entity pairs.
(2) 
Existing FKGC methods generate the representations of few-shot relations based on the semantic information of entity pairs, without considering of the relation associated by the head and tail entities.
In practice, in real-world KGs, there may exist multiple relations between a pair of entities.
Modeling only the semantic information of head and tail entities cannot accurately infer the concrete relationship between the two entities in a specific situation.

To address the above limitations,
in this paper, we propose a Relation-Specific Context Learning (RSCL) framework for few-shot knowledge graph completion.
The RSCL framework is as follows.
Firstly, we present a subgraph extractor to construct a context subgraph for each triple, which contains multiple entity-relation paths surrounding the triple.
Secondly, we devise a hierarchical attention network, in which a Transformer-based global encoder and a graph neural networks-based local encoder are jointly trained to model the graph contexts of triples.
Instead of learning entity-pair representations, the hierarchical attention network generates triple representations that incorporate the semantic information of relations and entities simultaneously.
Finally, we design a hybrid attention aggregator to learn relation-specific representations for few-shot relations and evaluate the likelihood of query triples at the global and local levels.

The main contributions of this paper can be summarized as follows:
\begin{itemize}

\item We propose a relation-specific context learning framework, which exploits graph contexts of triples to learn relation-specific relations that incorporate the semantic information of relations and entities simultaneously.

\item We design a novel hierarchical attention network, which combines the advantages of Transformer and graph neural networks to capture rich graph contextualized information of triples and highlight valuable local neighborhood information of head and tail entities.

\item We conduct extensive experiments on two public datasets.
Experimental results demonstrate that our method outperforms the state-of-the-art methods of few-shot KG completion.
Additionally, the ablation study verifies the effectiveness of each module in our framework.

\end{itemize}

\section{Related Work}
\label{RW}
\subsection{Traditional Knowledge Graph Completion}
The knowledge graph completion (KGC) task aims to infer new triples by using existing ones in KGs.
A popular approach for KGC is to embed entities and relations into low-dimensional distributed representations, and then use a well-designed score function to measure the plausibility of query triples.
Existing knowledge graph embedding methods are broadly classified into two main branches: translation models and semantic matching models.

Translation models interpret relations as translation operations between entity pairs and define a distance-based score function accordingly.
TransE \cite{bordes2013translating} is the seminal work of translation models, which represents each relation as a translation vector connecting the head and tail entities.
Despite its simplicity and efficiency, TransE cannot deal with 1-to-many, many-to-1, and many-to-many complex relations.
Subsequently, TransH \cite{wang2014knowledge} introduces relation-specific hyper-planes to endow different roles of an entity with different representations.
TransR \cite{lin2015learning} extends TransH and projects entities and relations into a separate entity space and relation spaces, respectively.

Semantic matching models exploit similarity-based score functions by matching latent semantics of entities and relations.
RESCAL \cite{nickel2011three} is a tensor factorization approach that associates each relation with a relation matrix and models the pairwise interactions between two entities.
DistMult \cite{Yang2015Embedding} simplifies RESCAL and restricts the relation matrices to be diagonal matrices, while the over-simplified DisMult cannot handle asymmetric relations. 
To model both symmetric and asymmetric relations, ComplEx \cite{trouillon2016complex} represents entities and relations as complex vectors instead of real-valued ones.
In addition to the linear/bilinear models listed above,
some neural network-based models have been developed in recent years.
These models encode semantic matching
using more complex neural network architectures, such as standard multilayer perceptron \cite{dong2014knowledge}, convolutional neural networks \cite{nguyen2018novel,dettmers2018convolutional}, 
and graph neural networks \cite{schlichtkrull2018modeling,shang2019end}.

Despite their success, 
these traditional KGC models usually assume the availability of sufficient training triples for all relations.
In contrast, our model focuses on performing the KG completion task in few-shot scenarios.
For example, there is currently a severe lack of research samples on the new virus, COVID-19.
In this realistic scenario, our model can effectively complete COVID-19 KGs using only limited training data.
This is of great significance for accelerating the research process of COVID-19 and facilitating downstream information retrieval tasks such as drug discovery\cite{mohamed2020discovering}.

\subsection{Few-shot Knowledge Graph Completion}
Up to now, 
there are two groups of FKGC models:
(1) Metric learning-based methods:
GMatching \cite{xiong2018one} is the first research on FKGC.
It leverages a neighbor encoder to learn entity embeddings with their direct neighbors, and a multi-step matching processor to measure the similarity between the query and reference entity pairs.
FSRL \cite{zhang2020few} extends GMatching from the one-shot scenario to the few-shot scenario, and introduces an attention mechanism to encode direct neighbors with different weights.
FAAN \cite{sheng2020adaptive} argues that entities should have various attributions in different task relations and learns entity representations adaptive to different relations.
PINT \cite{xu2021p} represents each entity pair with the paths from the head to tail entities, and then measures the degree of semantic matching between the reference and query entity pairs by computing their path interactions.
(2) Meta learner-based methods: 
MetaR \cite{chen2019meta} focuses on transferring relation-specific information so that a new, unseen relation can be rapidly learned after several gradient descent updates.
To this end, it utilizes a relation-meta learner to generate relation information from the embeddings of entity pairs.
GANA-MTransH \cite{niu2021relational} shares a similar idea with MetaR, but learns the relation-specific hyper-plane parameters to model complex relations.

Our work differs from existing FKGC methods in two aspects.
First, previous methods learn entity-pair representations from the local neighborhoods of head and tail entities, ignoring that a large number of entities are actually long-tailed and have a few direct neighbors.
Our method learns global-aware and local-aware triple representations by encoding the contextualized information of triples.
Second, previous methods exploit the semantic information of head and tail entities to learn representations of few-shot relations, while our work generates relation representations by incorporating the representations of head entities, relations, and tail entities.


Graph contexts of triples have also been explored in other research fields outside FKGC to improve entity embeddings \cite{goikoetxea2015random,luo2015context,wang2019coke,wang2019capturing}.
CoKE \cite{wang2019coke} encodes the graph contexts of entities and relations via BERT \cite{devlin2019bert} to learn
contextual KG embeddings.
RSN \cite{guo2019learning} utilizes recurrent skipping networks equipping with a skipping mechanism to model the long-term relational paths.
Similarly, RW-LMLM \cite{wang2019capturing} adopts the transformer
blocks to encode the entity-relation paths sampled by a random walk strategy.
All these studies demonstrate the efficacy of modeling graph contexts when learning KG embeddings.



%

\section{Background and Overview}
\subsection{Problem Setting}
Knowledge graph $\mathcal{G}$ is a collection of factual knowledge.
Each fact is organized in the form of a triple $\{(h,r,t)\} \subseteq \mathcal{E} \times \mathcal{R} \times \mathcal{E}$ where $\mathcal{E}$ and $\mathcal{R}$ are the entity set and relation set, respectively.
Given arbitrary two of three elements within a triple, the KG completion task aims to predict the remaining one.
This work focuses on predicting $t$ given $h$ and $r$ since our goal is to infer new facts for few-shot relations.
Unlike previous studies that usually assume sufficient training triples are available, this work studies the practical scenario where only a limited number of training triples are given, i.e., few-shot KG completion.
The task of FKGC is to predict $t$ in the query triple $(h,r,?)$ by observing $K$ reference triples of a few-shot relation $r$.

\subsection{Overview of FKGC}

Formally, a KG can be represented as  $\mathcal{G}=\{(h,r,t)\in \mathcal{E} \times \mathcal{R} \times \mathcal{E} \}$, where each relation $r \in \mathcal{R}$ is viewed as a task.
A background KG $\mathcal{G'}$ is represented as a subset of $\mathcal{G}$ with some tasks from $\mathcal{R}$. 
The remaining tasks in $\mathcal{R}$ are further divided into three disjoint sets $\mathcal{R}_{train}$, $\mathcal{R}_{test}$ and $\mathcal{R}_{valid}$.
The three task sets are used in the meta-training, meta-testing, and meta-validation phases, respectively.

In the meta-training phase, for each training task $r\in \mathcal{R}_{train}$, its related triples are randomly split into a support set $\mathcal{S}_r$ and a query set $\mathcal{Q}_r$.
The $\mathcal{S}_r=\{(h_i,r,t_i)\}^{|K|}_{i=1}$ consists of $K$ reference triples.
The $\mathcal{Q}_r=\{(h_j,r,t_j,\mathcal{C}_{h_j,r}) \}^{|\mathcal{Q}_r|}_{j=1}$ contains $|\mathcal{Q}_r|$ triples that need to be predicted,
where $t_j$ is the ground-truth tail entity for each query triple $(h_j,r,?)$ and $\mathcal{C}_{h_j,r}$ is the set of the corresponding candidate entities in $\mathcal{G}$.
The candidate entities are constructed based on the entity type constraint \cite{xiong2018one,toutanova2015representing}.
In the meta-training phase, the FKGC model is trained using $K$ labeled triples in $\mathcal{S}_r$ so that it can rank the ground-truth entity $t_j$ higher than the corresponding candidate entities in $\mathcal{C}_{h_j,r}$.

After sufficient training with $\mathcal{R}_{train}$, the learned model can be used for the meta-validation and meta-testing phases with $\mathcal{R}_{valid}$ and $\mathcal{R}_{test}$.
The two phases are the same, and we only describe the meta-testing phase.
In the meta-testing phase,
each test relation $r'\in \mathcal{R}_{test}$ has also its own support set $\mathcal{S}_{r'}$ and query set $\mathcal{Q}_{r'}$, defined in the same way as training tasks.
To predict a given triple $(h,r,?) \in \mathcal{Q}_{r'}$, the learned model is utilized to compute the similarity score between the $K$ references in $\mathcal{S}_{r'}$ and each candidate triple.  
The candidate entity corresponding to the highest score is deemed as the missing tail entity of $(h,r,?)$.
The above procedure is repeated for all test tasks in $\mathcal{R}_{test}$.

\section{Our Approach}
\begin{figure*}[!t]
\centering	
\subfigure{\includegraphics[width=0.95\linewidth, trim=0 0 0 0, clip]{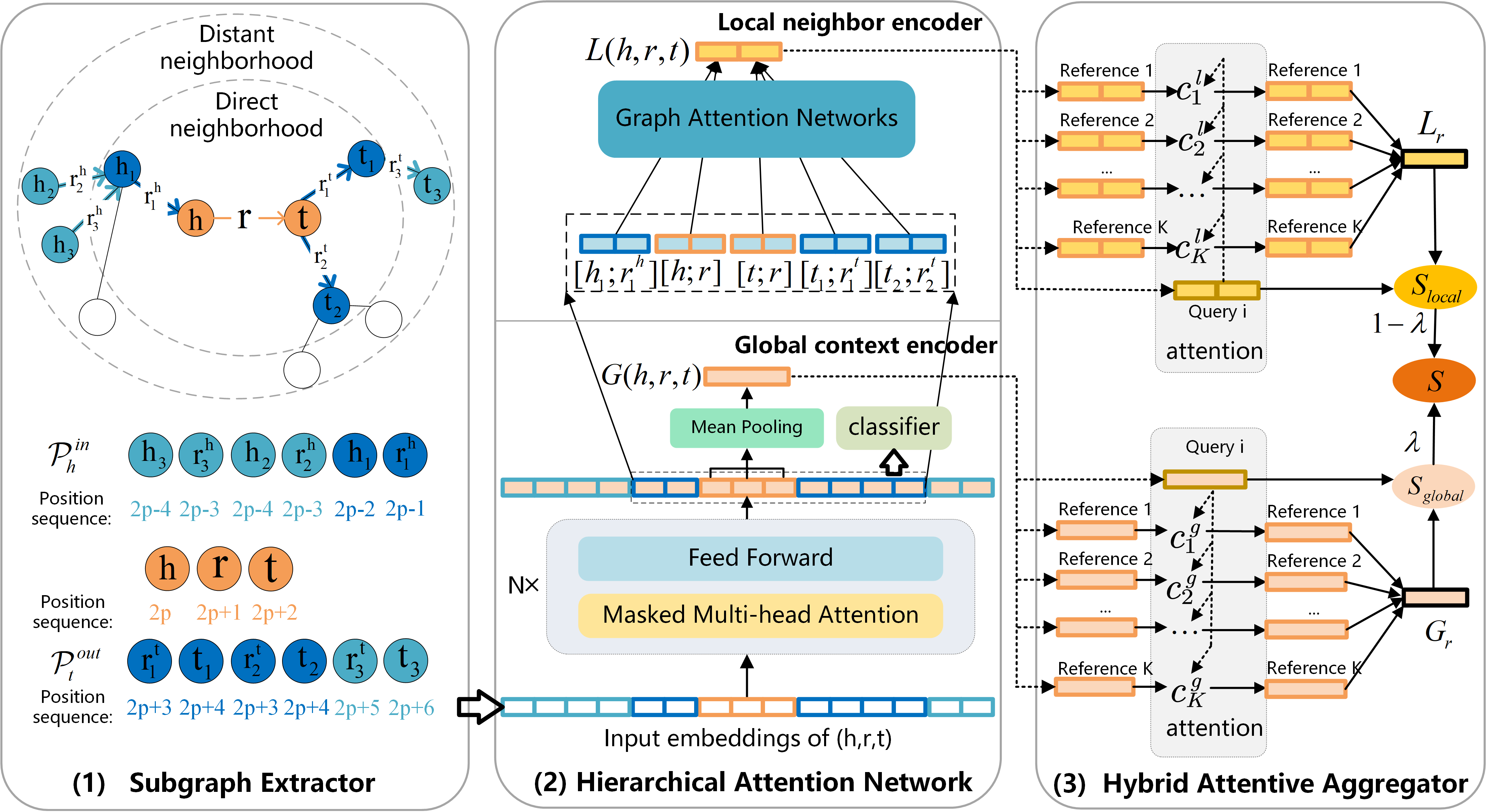} }
\caption{
	An overview of the proposed RSCL. We first extract the graph context for a given triple via (1) subgraph extractor. In our case, three neighboring entity-relation pairs are selected for each entity (i.e., $p=3$), and the maximum of distant neighbors $q$ is set to $2$.
	Then, we learn global-aware and local-aware representations for triples in the support and query sets via (2) hierarchical attention network.
	Finally, (3) hybrid attentive aggregator aggregates the global-aware representations of reference triples into a global relation-specific representation $G_r$, and the local-aware triple representations into a local relation-specific representation $L_r$.
	The similarity scores $S_{global}$ and $S_{local}$ are combined to measure the plausibility of the query triple.
}
\label{HTGNNet}
\end{figure*}

This section presents our framework RSCL.
The purpose of RSCL is to learn a metric function, which evaluates the likelihood of the input query based on the similarity score between the query and the support references.
To achieve this goal, RSCL consists of three modules: 
(1) subgraph extractor to construct the graph contexts of triples from background graph $\mathcal{G}'$;
(2) hierarchical attention network to generate global-aware and local-aware representations of triples by encoding the extracted graph contexts;
(3) hybrid attentive aggregator to learn global and local relation-specific representations and compare the query and the support references.
Figure \ref{HTGNNet} shows the overall architecture of RSCL.
Next, we will detail each module of RSCL via one few-shot task $r \in \mathcal{R}_{train}$, and its corresponding support set $\mathcal{S}_r$ and query set $\mathcal{Q}_r$.

\subsection{Subgraph Extractor}


\begin{figure*}[!t]
\centering	
\subfigure{\includegraphics[width=0.95\linewidth, trim=0 0 0 0, clip]{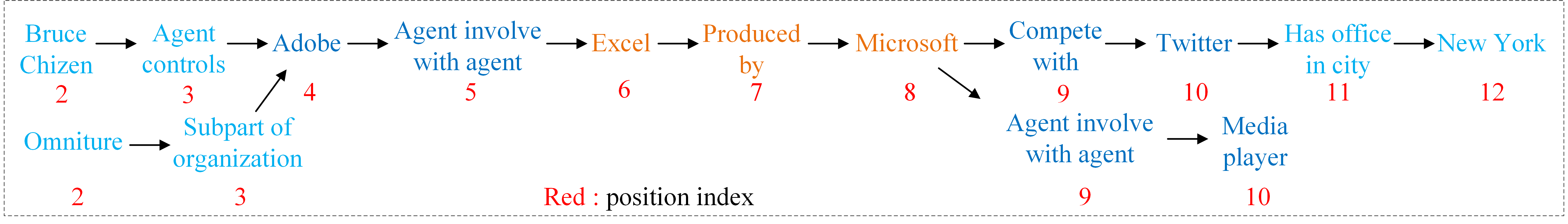}}
\caption{Example of our proposed position encoding.
	The target triple is (\texttt{Excel}, \texttt{Produced by}, \texttt{Microsoft}).
	The head entity \texttt{Excel} has one direct neighbor and two distant neighbors.
	The tail entity \texttt{Microsoft} has two direct neighbors and one distant neighbor. 
	The position indices of the entity-relation sequences are as follows: \{2, 3, 2, 3, 4, 5, 6, 7, 8, 9, 10, 9, 10, 11, 12\}.
}
\label{pos}
\end{figure*}

To obtain the contextualized information of triples, we propose 
a subgraph extractor to generate a subgraph composed of the direct and distant neighbors of the head and tail entities.

In specific, given a triple $(h,r,t)$ about a few-shot relation $r$, its graph context is defined as an entity-relation subgraph $\mathcal{P}_{(h,r,t)}$ surrounding the target triple, i.e., $\mathcal{P}_{(h,r,t)}=\mathcal{P}^{in}_h \cup (h,r,t) \cup \mathcal{P}^{out}_t$, where $\mathcal{P}^{in}_h$ is the in-degree entity-relation subgraph of the head entity $h$, and $\mathcal{P}^{out}_t$ is the out-degree entity-relation subgraph of the tail entity $t$.
We obtain the in-degree entity-relation subgraph $\mathcal{P}^{in}_h$ as follows:
\begin{enumerate}
\item Extract the direct in-degree entity-relation pairs $\mathcal{N}^{dre}_h$ for the head entity $h$:
\begin{equation}
	\mathcal{N}^{dre}_h = \{(e^{h}_i,r^{h}_i)|(e^{h}_i,r^{h}_i,h)\in \mathcal{G}'\}^{|\mathcal{N}^{dre}_h|}_{i=1},
\end{equation}
where $|\mathcal{N}^{dre}_h|$ denotes the number of direct neighbors of $h$.
If there are multiple relations between $h$ and $e^h_i$, we will traverse all relations.
\item Extract the distant in-degree entity-relation pairs $\mathcal{N}^{dst}_h$ for the head entity $h$:
\begin{equation}
	\mathcal{N}^{dst}_h = \{(e^{h}_i,r^{h}_i)|(e^{h}_i,r^{h}_i,h')\in \mathcal{G}', h'\in\mathcal{E}'_h\}^{|\mathcal{N}^{dst}_h|}_{i=1},
\end{equation}
where $\mathcal{E}'_h$ denotes the set of directly neighboring entities of $h$, and $|\mathcal{N}^{dst}_h|$ denotes the number of 
distant neighbors of $h$.
To be generalized to large-scale
KGs, our subgraph extractor only expands distant neighbors to second-order neighbors.
We also traverse all the relations between $h'$ and $e^h_i$.
\item Generate the in-degree entity-relation subgraph $\mathcal{P}^{in}_h$ for the head entity $h$:
\begin{equation}
	\mathcal{P}^{in}_h= \{(e^h_i,r^h_i)^{dre}\}^{|\mathcal{N}^{dre}_h|}_{i=1} \cup \{(e^h_i,r^h_i)^{dst}\}^{q}_{i=1},
\end{equation} 
where an entity-relation pair $(e^h_i, r^h_i)^{dre}$ is randomly chosen from the direct neighborhood $\mathcal{N}^{dre}_h$, and $(e^h_i, r^h_i)^{dst}$ from the distant neighborhood $\mathcal{N}^{dst}_h$.
This random selection ensures that all neighbors of an entity can be sampled.
$q$ is the maximum that we restrict the number of distant neighbors to since excessive distant neighbors may introduce additional noises.
\end{enumerate}

For the tail entity $t$, we conduct a similar process to extract its direct and distant out-degree entity-relation pairs $\mathcal{N}^{dre}_t$ and $\mathcal{N}^{dst}_t$, and generate its out-degree entity-relation subgraph $\mathcal{P}^{out}_t=\{(r^t_i,e^t_i)^{dre}\}^{|\mathcal{N}^{dre}_t|}_{i=1} \cup \{(r^t_i,e^t_i)^{dst}\}^{q}_{i=1}$.

In this work, we select $p$ neighboring entity-relation pairs for each entity so that the size of the entity-relation subgraph $\mathcal{P}_{(h,r,t)}$ can be fixed to $4p+3$.
In addition, we use a special token (\texttt{[PAD]}) to fill the final subgraph $\mathcal{P}_{(h,r,t)}$ when its size is less than $4p+3$.
The filled subgraph is considered as the graph context of triple $(h,r,t)$.
Note that, during the stages of meat-testing and meta-validation, we do not randomly choose entity-relation pairs and extract only top $p$ ones for each entity. 
Figure \ref{pos} shows an instance of the extracted entity-relation subgraph.
We assign a position index to each token in the set, which will be described in the next section.

\subsection{Hierarchical Attention Network}
In this subsection, we present a hierarchical attention network to learn triple representations by encoding the extracted subgraphs of triples.
The proposed hierarchical attention network consists of two components: a global context encoder to generate global-aware representations of triples by modeling their contextualized information, and a local neighbor encoder to learn local-aware representations of triples by aggregating the direct neighborhoods of their head and tail entities.

\subsubsection{Global context encoder}
A conventional choice for modeling graph-structured data is graph neural networks.
However, using graph neural networks can only learn the representations of head and tail entities in triples, but cannot acquire the relation representations that are crucial for inferring the underlying relational information between two entities.
To solve this problem, we model graph contexts of triples as entity-relation sequences and develop a graph context encoder based on Transformer to acquire the representations of both entities and relations, simultaneously.
Although Transformer \cite{vaswani2017attention} has proven to be powerful in modeling sequential data, directly applying the vanilla Transformer for encoding the entity-relation sequences is sub-optimal as the self-attention used in Transformer cannot model the structural information of graph contexts.
In this paper, we introduce a simple but effective position encoding to capture the node structural information reflected on the neighboring entity-relation pairs.


In specific, with the generated graph context $\mathcal{P}_{(h,r,t)}$ of triple $(h,r,t)$, we first construct its input embedding.
For a clear description, we represent $\mathcal{P}_{(h,r,t)}$ as a token sequence $\mathcal{X}=\{x_1,\dots,x_{(4p+3)}\}$,
where elements with odd indices represent entities and the remaining ones are intermediate relations.
For each element $x_i$ in $\mathcal{X}$, its input embedding $z_i$ is the sum of element embedding and position embedding:
\begin{equation}
z_i = x^{\text{ele}}_i + x^{\text{pos}}_i,
\end{equation}
where $x^{\text{ele}}_i$ is the element embedding, $x^{\text{pos}}_i$ is the position embedding.
For position embedding, we assign each token (entity or relation) a position index to characterize the graph structure of the entity-relation sequence.
We set the position indices of head entity $h$, relation $r$ and tail entity $t$ as $2p$, $2p+1$, and $2p+2$, respectively.
For the head entity, the position indices of its $k$-hop neighboring entity and relation are $2p-k-1$ and $2p-k$, respectively.
We assign the position indices of the tail entity in a similar way.
Figure \ref{pos} illustrates an example of our position encoding.
As we can see, the direct neighboring entities of \texttt{Microsoft} (i.e., \texttt{Twitter} and \texttt{Media player}) are assigned the same position index, and thus their semantic information can be aggregated equally by our global context encoder.
Different from previous FKGC methods \cite{xiong2018one,zhang2020few,sheng2020adaptive} that utilize the ready-made KG embedding models to pre-train element embeddings, we initialize the element and position embeddings randomly and then update them during training.
Thanks to the powerful ability of Transformer in modeling the long-term dependencies, the updated entity and relation embeddings can incorporate more contextualized information.

The transformer block consists of two main parts: masked multi-head attention layer and feed forward layer.
We use $L$ stacked transformer blocks to encode the input representation of $\mathcal{X}$ and obtain:
\begin{equation}
z^l_i={\rm Transformer}(z^{l-1}_i),\quad l\in [1,L],
\end{equation}
where $z^l_i$ is the hidden state of $x_i$ after the $l$-th layer.
Considering that the use of Transformer has become ubiquitous in recent years,
we omit a comprehensive description and refer readers to \cite{vaswani2017attention} and \cite{devlin2019bert}.

The the semantic representations of head entity $h$, tail entity $t$, and relation $r$ are the final hidden states of the corresponding positions, respectively, i.e., $z^L_{2p}$, $z^L_{2p+2}$, and $z^L_{2p+1}$.
In order to avoid the high complexity, we employ a mean pooling layer to incorporate their semantic representations and generate the global-aware representation of the triple $(h,r,t)$, as shown below:
\begin{equation}
	G{(h,r,t)}= {\rm MeanPooling}(z^L_{2p},z^L_{2p+1},z^L_{2p+2})\\
	\label{mp},
\end{equation}



\subsubsection{Local neighbor encoder}
The direct neighbors of an entity are the most important neighborhood to characterize the semantic information of the entity \cite{sun2020knowledge}.
Based on this idea, we utilize graph attention networks \cite{velivckovic2018graph} as the backbone model and devise a local neighbor encoder to learn local-aware triple representations by aggregating the local neighborhood information of head and tail entities.
In fact, other variants of graph neural networks are also feasible.
We leave the investigation of other graph neural networks \cite{kipf2016semi,schlichtkrull2018modeling} to further work.

For the given triple $(h,r,t)$, we can access its direct neighborhoods $\mathcal{N}^{dre}_h$ and $\mathcal{N}^{dre}_{t}$.
For convenience of description, we generalize both the head entity $h$ and the tail entity $t$ as entity $e$.
For each direct neighbor $(e_i,r_i)\in N^{dre}_e$ of entity $e$,
we first calculate its attention weight $\alpha_i$ as follows:
\begin{gather}
d_i = W_1[\boldsymbol{e_i};\boldsymbol{r_i}],\\
\alpha_i = \frac{{\rm exp} \big({\rm LeakyReLU}(U^T_1d_i)\big)}{\sum_{(e_j,r_j)\in \mathcal{N}^{dre}_e}{\rm exp} \big({\rm LeakyReLU}(U^T_1d_j)\big)},
\label{attention}
\end{gather}
where $\boldsymbol{e_i} \in \mathbb{R}^d$ and $\boldsymbol{r_i}\in \mathbb{R}^d$ are the semantics representations of $e_i$ and $r_i$ learned by the global context encoder;
$d_i$ is the representation of neighbor $(e_i,r_i)$;
$W_1 \in \mathbb{R}^{2d\times 2d}$ is a linear transformer matrix;
$U_1\in \mathbb{R}^{2d}$ denotes a weight vector followed by a $\rm{LeakyReLU}$ activation function.

Then we compute the weighted neighbor representation of $e$ by combining the attention weights and the neighbor representation of $e$, as shown below:
\begin{equation}
	\boldsymbol{e_{nbr}} = \sum_{(e_i,r_i)\in N^{dre}_e}\alpha_id_i
\end{equation}
Finally, we obtain the entity representation of $e$ by coupling the embedding of $e$ and its weighted neighbor representation, as shown below:
\begin{gather}
\boldsymbol{e'} = \boldsymbol{e}+W_2\boldsymbol{e_{nbr}},
\label{eq9}
\end{gather}
where $W_2\in \mathbb{R}^{2d\times d}$ is a learnable weight vector.


We perform the above procedures for the head entity $h$ and the tail entity $t$ and obtain their entity representations $\boldsymbol{h'}$ and $\boldsymbol{t'}$.
We further combine them with the embedding of relation $r$ to obtain the local-aware representation of triple $(h,r,t)$.
The calculation is as follows:
\begin{equation}
L{(h,r,t)} = {\rm {LayerNorm}}((\boldsymbol{h'}+W_r\boldsymbol{r})\oplus(\boldsymbol{t'}+ W_r \boldsymbol{r})),
\end{equation}
where ${\rm {LayerNorm}}$ denotes layer normalization \cite{ba2016layer}; $W_r\in \mathbb{R}^{d\times d}$ is a learnable weight vector; $\oplus$ is the concatenation operation; $\boldsymbol{r}$ is the semantic representation of relation $r$ learned by the global context encoder.
Note that the local neighbor encoder only uses the direct neighbors in the entity-relation paths, instead of all of the direct neighbors of entities in KGs, which allows the global and local encoder to be updated synchronously.



\subsection{Hybrid Attentive Aggregator}
To make predictions, we propose a hybrid attentive aggregator to learn global and local relation-specific representations by aggregating the representations of reference triples.
Intuitively, references with more similar features to the query are more conductive to making accurate predictions for the query.
Based on this idea, we employ an attention mechanism to assign different weights to reference triples and make the model to focus more on the query-relevant references.


Specifically, given a query triple $q_r$, we compute the global relation-specific representation $G_r$ of few-shot relation $r$ by aggregating the global-aware representations of reference triples with different weights.
The calculation is as follows:
\begin{gather}
G_r = \sum_{s_{ri}\in \mathcal{S}_r}c^g_i G(s_{ri}),\\
c^g_i = \frac{{\rm exp} (G(q_r) \odot G(s_{ri}))}{\sum_{s_{rj} \in \mathcal{S}_r}{\rm exp}(G(q_r) \odot G(s_{rj}))}.
\end{gather}
Here, $s_{ri} \triangleq (h_i,r,t_i)\in \mathcal{S}_r$ represents the $i$-th reference triple in the support set $\mathcal{S}_r$; $G(s_{ri})$ is the global-aware representation of triple $s_{ri}$; $\odot$ denotes element-wise production.
Meanwhile, we compute the local relation-specific representation $L_r$ of $r$ in a similar way:
\begin{gather}
	L_r = \sum_{s_{ri}\in \mathcal{S}_r}c^l_i L(s_{ri})\\
	c^l_i = \frac{{\rm exp} \big(L(q_r) \odot L(s_{ri})\big)}{\sum_{s_{rj} \in \mathcal{S}_r}{\rm exp}\big(L(q_r) \odot L(s_{rj}) \big)}
\end{gather}

With the generated global and local relation-specific representations, we compare the given query $q_r$ and the reference triples at both global and local levels.
At the global level, we calculate the similarity score between the global-aware representation of $q_r$ and the global relation-specific representation of $r$, as shown below:
\begin{equation}
	S_{global} =  G_r \odot G(q_r).
\end{equation}
At the local level, we compute the similarity score between the local-aware representation of $q_r$ and the local relation-specific representation, as shown below:
\begin{equation}
	S_{local} =  L_r \odot L(q_r).
\end{equation}

Finally, we utilize the linear interpolation of $S_{global}$ and $S_{local}$ to measure the plausibility of the query triple $q_r$.
The final similarity score between the reference triples and the query can be obtain:
\begin{equation}
S = \lambda S_{global} + (1-\lambda)S_{local},
\label{eq_lambda}
\end{equation}
where hyper-parameter $\lambda$ controls the tradeoff between the global similarity score and the local similarity score. 
$\lambda=1$ means that we estimate the plausibility of query triples at the global level, and $\lambda=0$ means that we only exploit the local relation-specific information to predict query triples.

\subsection{Loss Function and Model Training}

Given a task $r\in \mathcal{R}_{train}$ and its corresponding triples, we randomly sample $K$ positive triples to construct the support set $\mathcal{S}_r$, and $B$ triples to construct the positive query set $\mathcal{Q}_r$, where $B$ is the batch size.
In addition, we also construct a set of negative query triples $\mathcal{Q}^-_r$ by polluting the tail entities of positive queries in $\mathcal{Q}_r$, i.e., $\mathcal{Q}^-_r=\{(h_q,r,t^-_q)\lvert (h_q,r,t_q) \in \mathcal{Q}_r, t^-_q \in \{\mathcal{C}_{h_q,r} \setminus t_q\} \}$.
With the reference triples, the positive queries and the negative queries, two similar scores $S^+$ and $S^-$
can be calculated via the proposed RSCL.
Finally, a ranking loss related to the two scores is minimized for performing few-shot knowledge graph completion:
\begin{equation}
\mathcal{L}_{ranking} =  max(0,\gamma+S^--S^+),
\end{equation}
where hyper-parameter $\gamma$ is a margin distance between the positive and negative queries.

In addition, to enhance the hierarchical attention network robustness, 
we employ the masking strategy used in BERT \cite{devlin2019bert} and randomly mask 15\% of tokens in the input sequence $\mathcal{X}$.
To predict the masked tokens, we introduce a cross-entropy loss as follows:
\begin{gather}
\mathcal{L}_{masking}=-\sum ^{4p+3}_{i=1} y_i\log(p_i),\\
p_i=softmax(z^L_iW_c),
\end{gather}
where $z^L_i$ is the final hidden state of the transformer blocks corresponding to $x_i$; $p_i$ is the predict label of $x_i$ obtained from the classifier;
$W_c$ is the classifier weight; $y_i$ is the true label of $x_i$

The final function loss is defined as follows:
\begin{equation}
\mathcal{L} = \mathcal{L}_{ranking} + \mathcal{L}_{masking}.
\end{equation}
Following \cite{zhang2020few}, we adopt a batch sampling based meta-training procedure to minimize $\mathcal{L}$ and optimize model parameters.

\section{Experiments}
In this section, we conduct extensive experiments on two public datasets to evaluate the performance of our method and verify the effectiveness of the key components in the proposed framework.

\begin{table}
\caption{The statistics of the two datasets: NELL-One and Wiki-One.}
\centering
\renewcommand\arraystretch{1.2}
\setlength{\tabcolsep}{1.3mm}{
	\begin{tabular}{c|c|c|c|c|c}
		\hline
		Dataset  &$|\mathcal{E}|$ &$|\mathcal{R}|$ &$|\mathcal{R}_{train}|$ &$|\mathcal{R}_{test}|$ &$|\mathcal{R}_{valid}|$\\
		\hline
		NELL-One &68,545 &358  &51 &11&5 \\
		Wiki-One  &4,838,244 & 822 &133 &34 &16 \\
		\hline
\end{tabular}}
\label{Dataset}
\end{table}

\subsection{Experiment Setup}
\subsubsection{Datasets}
In our experiments, we choose two widely used benchmark datasets, namely NELL-One and Wiki-One\footnote{https://github.com/xwhan/One-shot-Relational-Learning.}, for few-shot KG completion \cite{xiong2018one}.
The NELL-One dataset is based on NELL \cite{mitchell2018never}, 
a system that continuously collects structured knowledge from webs via an intelligent agent.
The Wiki-One dataset is based on Wikidata\cite{vrandevcic2014wikidata}, a free structured knowledge base that is created and maintained by human editors.
Note that Wiki-One is an order of magnitude larger than any other datasets in terms of the numbers of entities and triples.
In both datasets, relations with less than 500 but more than 50 triples\footnote{Relations in the meta-testing set need enough triples for evaluation.} are selected as few-shot relations, and the remaining relations along with their triples constitute the background knowledge graphs.
There are 67 and 183 few-shot relations on NELL-One and Wiki-One, respectively. 
Following the original settings \cite{xiong2018one,zhang2020few,chen2019meta}, we 
split the training/test/validation relations as 51/11/5 and 133/34/16 on NELL-One and Wiki-One, respectively.
The detailed statistics of the two datasets are summarized in Table \ref{Dataset}, where $|\mathcal{R}_{train}|$, $|\mathcal{R}_{test}|$ and $|\mathcal{R}_{valid}|$ denote the number of relations used in the meta-training, meta-testing and meta-validation phases, respectively.

Traditional KG completion work typically utilizes standard benchmarks, such as FB15k \cite{bordes2013translating} and FB15k-237 \cite{toutanova2015representing}, to evaluate model performance.	
We do not use these datasets for few-shot KG completion because they often contain sufficient training triples for each relation, and consider the same set of relations during training and testing.
%



\subsubsection{Metrics}
To evaluate the model performance, we use two common metrics:
Mean Reciprocal Rank (MRR, the average
of the reciprocal ranks of the correct entities) and Hits@$N$ (the proportion of correct entities ranked in the top $N$).
For the two metrics, higher values of MRR or Hits@$N$ are indicative of better performance of KG completion.
The $N$ is set to $1$, $5$ and $10$.
The few-shot
size is set to 1 and 5 for the 1-shot and 5-shot KG completion tasks.

\subsubsection{Baselines}
We compare our model with two types of baseline models:
\begin{itemize}
\item Traditional embedding-based models. This type of model requires sufficient training triples to learn KG embeddings.
In our experiments, we adopt six widely used models:
TransE \cite{bordes2013translating},
TransH \cite{wang2014knowledge},
TransR \cite{lin2015learning},
RESCAL \cite{nickel2011three},
DisMult \cite{bordes2014semantic},
and ComplEx \cite{trouillon2016complex}.
To implement FKGC for these traditional methods, we utilize not only all the triples in the background KG $\mathcal{G'}$ and the training tasks, but also the reference triples from the support sets in the validation and test tasks to train them.
Then, we use the triples from the query sets in the validation and test tasks to evaluate the trained models.
We reproduce the experimental results of all these traditional models using OpenKE\footnote{https://github.com/thunlp/OpenKE/tree/OpenKE-PyTorch.} \cite{han2018openke} with their best hyperparameter settings reported in the original papers.
\item Existing FKGC models. This type of model\footnote{For the sake of fairness, we do not compare our method with P-INT \cite{xu2021p} because its results are obtained by discarding disconnected entity pairs.} achieves the best results on the FKGC task, including GMatching \cite{xiong2018one}, FSRL \cite{zhang2020few}, FAAN \cite{sheng2020adaptive}, MetaR \cite{chen2019meta}, and GANA \cite{niu2021relational}.
Among them, GMatching, FSRL, FAAN, and GANA enhance entity representations by encoding their direct neighbors and generate the representations of few-shot relations using the enhanced representation of head and tail entities in the reference triples.
MetaR extracts entity-pair specific relation meta by modeling the embeddings of head and tail entities,
and then computes relation representations by averaging all entity-pair relation meta in the reference triples. 
\end{itemize}

\subsubsection{Implementation Details}

We select the optimal hyperparameters of the proposed method by grid search on the validation set.
For a fair comparison, following GMatching \cite{xiong2018one}, we set the 
embedding dimensions to 100 for NELL-One and 50 for Wiki-One.
We also set the layer number of the transformer blocks to 4 for NELL-One and to 2 for Wiki-One, and each layer contains 4 heads.
To avoid overfitting, we apply
dropout to the global context encoder and the local neighbor encoder with the rate tuned in \{0.3,0.2\}.
The maximum number of neighbors $p$ is set to $8$, and the maximum number of distant neighbors $q$ is set to $5$.
The tradeoff parameter $\lambda$ is $0.4$.
The margin $\gamma$ is fixed to $5.0$.
During training, we apply mini-batch gradient descent to update the network parameters of our model.
The batch size is $256$ for NELL-One and $64$ for Wiki-One, respectively.
Moreover, we use the
Adam optimizer \cite{kingma2014adam} as the optimizer with an initial learning rate of $10^{-3}$, and the
weight decay is 0.25 for each $10k$ training steps.
Following \cite{sheng2020adaptive}, we evaluate our model on the validation set at every $10k$ training steps, and save the best model when MRR reaches the highest value within $300k$ steps.
To speed up training, we pre-train our model with the masking loss function $\mathcal{L}_{masking}$ on the background graph $\mathcal{G'}$ and the training tasks $\mathcal{R}_{train}$ before training begins.
The pre-trained embeddings are then used to initialize the element and position embeddings of entities and relations.
We employ Pytorch to implement our method and further conduct it on a server with 2 NVIDIA GeForce RTX 3090 GPUs.

\begin{table*}[!tbp]
	\caption{Evaluation results of all methods on NELL-One and Wiki-One in terms of MRR and Hits@\{1, 5, 10\}. For each metric, the best results are boldfaced, and the underlined numbers are the second best results.}
	\centering
	\renewcommand\arraystretch{1.2}
	\setlength{\tabcolsep}{2.8mm}{
		\begin{tabular}{ccc|cc|cc|cc}
			\hline
			& \multicolumn{2}{c}{MRR}  &\multicolumn{2}{c}{Hits@10}  &\multicolumn{2}{c}{Hits@5}  & \multicolumn{2}{c}{Hits@1}\\
			\cline{2-9}
			NELL-One	&1-shot &5-shot &1-shot &5-shot &1-shot &5-shot &1-shot &5-shot\\
			\hline
			
			TransE &0.062 &0.174 &0.169 &0.313 &0.072 &0.204 &0.004 &0.083 \\
			TransH &0.075 &0.159 &0.180&0.298 &0.085&0.183 &0.020&0.084\\
			TransR &0.062 &0.164 &0.174&0.333 &0.069&0.183 &0.006&0.083\\
			RESCAL &0.141 &0.187 &0.215 &0.341 &0.169&0.211 &0.092 &0.108 \\
			DisMult &0.102 &0.200 &0.177 &0.311 &0.126&0.251  &0.066 &0.137 \\
			ComplEx &0.131 &0.184  &0.233 &0.297 &0.086 &0.229 &0.086 &0.118 \\ 
			\hline
			GMatching &0.185 & 0.201 &0.313 &0.311 &0.260 &0.264 &0.119 &0.143 \\
			MetaR (In-Train) &0.250 & 0.261  &\underline{0.401} &0.437 &\underline{0.336} &0.350 &\underline{0.170} &0.168 \\
			MetaR (Pre-Train) &0.164 &0.209  &0.331 &0.355 &0.238 &0.280 &0.093 &0.141\\
			FSRL &0.192 &0.184 &0.326 &0.272 &0.262 &0.234 &0.119 &0.136 \\
			FAAN &0.198 & 0.279 &0.340 &0.428 &0.273 &0.364 &0.125 &\underline{0.200}\\
			GANA &\underline{0.254} &\underline{0.294} &\textbf{0.408} &\textbf{0.472} &0.335 &\underline{0.366} &0.164 &0.177\\
			\hline
			RSCL &\textbf{0.262} &\textbf{0.317}  &\underline{0.401}&\underline{0.452} &\textbf{0.342} &\textbf{0.386} &\textbf{0.186} &\textbf{0.243} \\								
			\hline\hline
			Wiki-One	&1-shot &5-shot &1-shot &5-shot &1-shot &5-shot &1-shot &5-shot\\	
			\hline	
			TransE &0.052 &0.083 &0.079 &0.117 &0.044 &0.075 &0.021 &0.052 \\
			TransH &0.086 &0.111 &0.151 &0.159 &0.080&0.114&0.043 &0.069\\
			TransR &0.058 &0.074 &0.074 &0.132 &0.049 &0.085 &0.011 &0.061\\
			RESCAL &0.077 &0.103 &0.081 &0.159 &0.074& 0.116&0.052 &0.086 \\
			DisMult &0.066&0.097  &0.108&0.161 &0.061 &0.119 &0.037 &0.044 \\
			ComplEx &0.075 &0.100 &0.125  &0.162 &0.059&0.122 &0.032&0.052 \\ 
			\hline
			GMatching &0.200 &0.263  &0.336 &0.387 &0.272&0.337& 0.120&0.197 \\
			MetaR (In-Train) &0.193 &0.221 &0.280 &0.302 &0.233 &0.264 &0.152 &0.178 \\
			MetaR (Pre-Train) &\textbf{0.314} &0.323  &\underline{0.404} &0.418 &\textbf{0.375} &\underline{0.385}&\textbf{0.266}  &0.270\\
			FSRL &0.082 &0.158 &0.173 &0.287 &0.105 &0.206 &0.029 &0.097 \\
			FAAN &0.226&\underline{0.341} &0.374&\underline{0.436} &0.302&0.395&0.153 &0.281\\
			GANA &0.203&0.283  &0.310&0.390 &0.258&0.339&0.146 &\underline{0.283}\\
			\hline
			RSCL &\underline{0.307}& \textbf{0.348}&\textbf{0.414 } & \textbf{0.454}&\underline{0.365} &\textbf{0.397} &\underline{0.241} &\textbf{0.289} \\	
			\hline
	\end{tabular}}
	\label{MainResult}
\end{table*}
\subsection{Comparison with FKGC Methods}

Table \ref{MainResult} reports the performance of all models on the NELL-One and Wiki-One datasets.
From the table, we can conclude that:
\begin{itemize}
\item On both datasets, our model significantly outperforms all the traditional KG embedding models.
The experimental results indicate that our method is more effective for knowledge graph completion in the few-shot scenario.

\item 
On both datasets, our model surpasses all the metric learning-based baselines (GMatching, FSRL, and FAAN).
Specifically, for 5-shot link prediction, our model achieves relative performance improvements of 3.8\% / 2.4\% / 2.2\% / 4.3\% in MRR / Hits@10 / Hits@5 / Hits@1 on NELL-One compared to the best performing FAAN \cite{sheng2020adaptive}.
These results demonstrate that exploiting graph contexts of triples can indeed improve the performance of few-shot KG completion.

\item Compared to MetaR that does not use the neighbor information of entities, our model outperforms it on NELL-One and obtains near-optimal results on Wiki-One.
For MetaR, only one of two settings (In-Train and Pre-Train) performs well on a single dataset.
Our model achieves good performance on both datasets, indicating that our method has better generalization ability for different datasets.

\item Compared to GANA\footnote{By retraining the GANA model, we observe that its results have a relatively large variance across runs. Hence, we report the average accuracy across multiple runs.}, our model outperforms it on Wiki-One in all metrics and obtains better performance in MRR, Hits@5, Hits@1 on NELL-One.
Since our model takes advantage of rich graph contexts of triples to represent few-shot relations, our model has more significant performance on the more sparse Wiki-One dataset.

\end{itemize}



\subsection{Ablation study and detail analysis}
\label{sec-abla}

Our framework is a joint learning framework including two main technologies: (1) modeling graph contexts of triples instead of using only direct neighbors of entities, and (2) devising a hierarchical attention network to learn global-aware and local-aware triple representations.
In this subsection, we perform an ablation study to better understand the function of each technology.
For simplicity, we report the results
of 5-shot references on NELL-One, as shown in
Table \ref{Ablat}:
\begin{itemize}
\item 
We investigate the effectiveness of modeling graph contexts by removing the distant neighbors from the entity-relation paths.
We observe a significant drop in
performance when distant neighbors are not used, which highlights the need for modeling graph contexts that incorporate distant neighbors.

\item We investigate the effectiveness of the hierarchical attention network by separately analyzing the effects of the global context encoder and the local neighbor encoder.
It will cause a significant performance drop if we remove any of the two encoders, which
indicates the large benefit of the hierarchical attention network to capture rich contextualized information and valuable neighbor information.
\end{itemize}

\begin{figure}[!t]
\centering	
\subfigure[Different $\lambda$]{\includegraphics[width=0.46\linewidth, trim=0 0 0 0, clip]{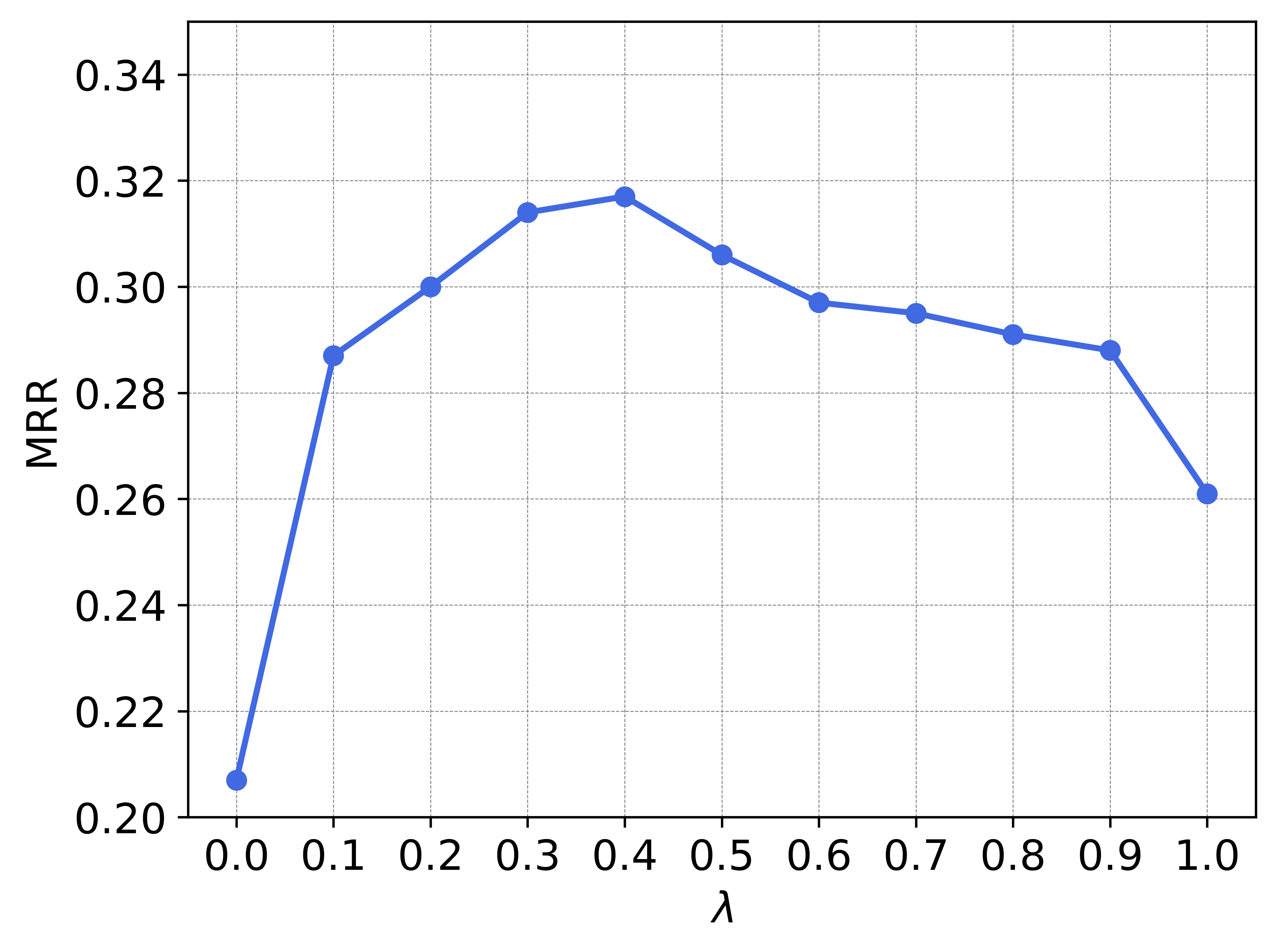}\label{la}}
\subfigure[Learning curves]{\includegraphics[width=0.48\linewidth, trim=0 0 0 0, clip]{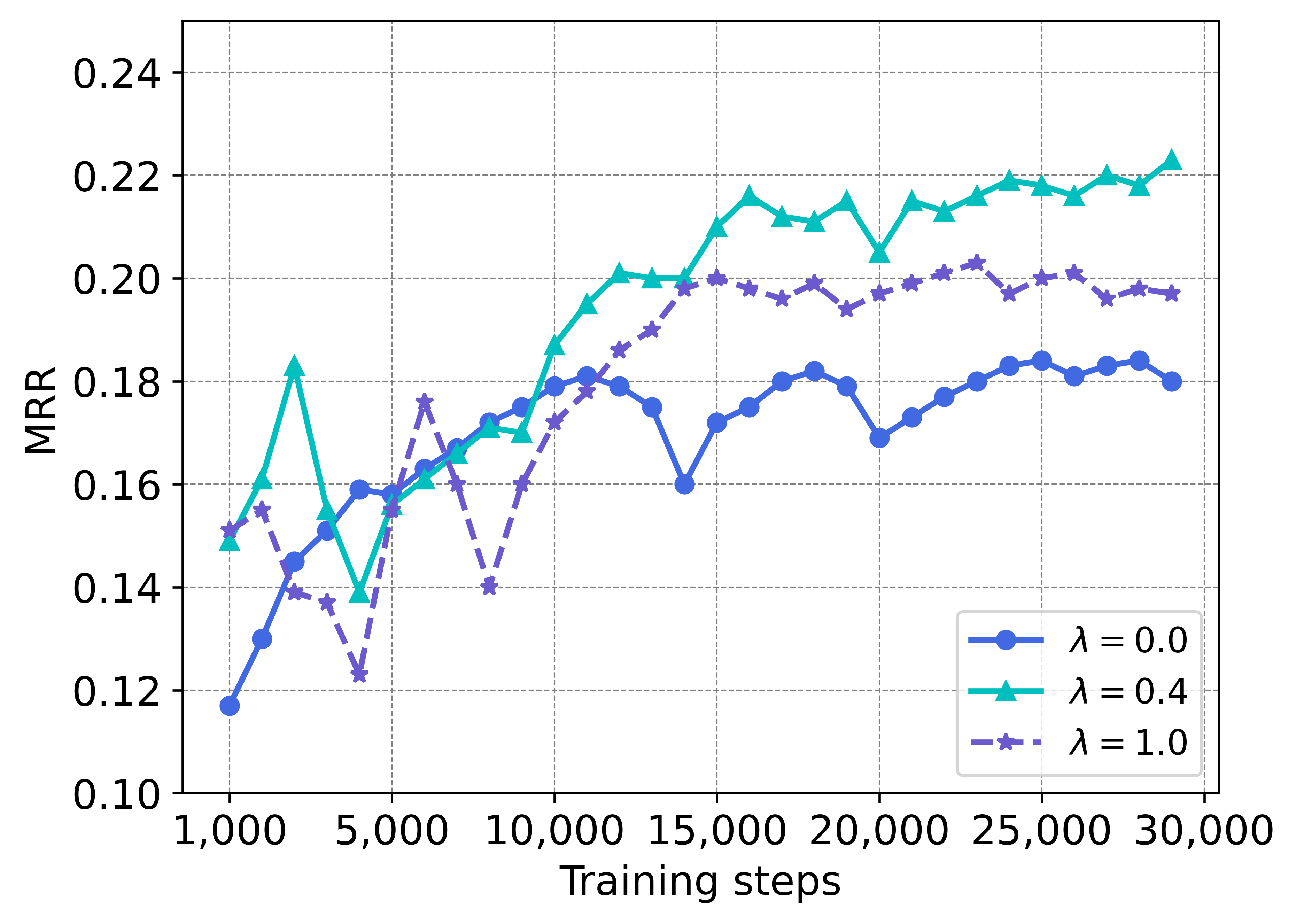}\label{lb}}
\caption{Illustrations of the learning process of RSCL with 5-shot references on NELL-One.
	The results of Figure \ref{la} and \ref{lb} are calculated on the test set and the validation set, respectively.
}
\end{figure}

To further explore the roles of the graph context encoder and the local neighbor encoder at each training step, we did a deeper analysis of the tradeoff parameter $\lambda$ used in Eq.\ref{eq_lambda}.
$\lambda=1.0$ means that only the graph context encoder is used, and $\lambda=0$ means that only the local neighbor encoder is used.
Figure \ref{la} reports the results of our method under different settings of $\lambda$ on NELL-One. 
This figure shows that:
(1) With the increase of $\lambda$, the performance of our method first increases and then decreases slightly, reaching a peak value at $\lambda=0.4$. 
It indicates that both the global context information and the local neighbor information should be combined to learn better triple representations.
(2) 
The MRR of $\lambda=1.0$ is higher than that of $\lambda=0$. 
This suggests that the graph context neighbor plays a more important role than the local neighbor encoder, further highlighting the necessity of modeling graph contexts of triples.
Figure \ref{lb} depicts the learning curves of $\lambda=0$, $\lambda=0.4$, and $\lambda=1.0$.
The curves are calculated on the validation set using the MRR metric.
We can see that at the beginning of model training (Training steps $<$ $10,000$), the MRR of $\lambda=1.0$ is generally lower than that of $\lambda=0$.
One potential reason could be that for some triples, there are some irrelevant distant neighbors introducing noisy information into the model.
As the number of training steps increases, the MRR of $\lambda=1.0$ is better than that of $\lambda=0$,
perhaps because our method effectively identifies the most informative neighbors via the hierarchical attention network.

\begin{table}[!tbp]
\caption{Ablation study on different components on NELL-One.}
\centering
\renewcommand\arraystretch{1.1}

\begin{tabular}{ccccc}
	\hline
	& \multicolumn{2}{c}{MRR} & \multicolumn{2}{c}{Hits@10} \\
	\cline{2-5}
	Configuration &1-shot &5-shot  &1-shot &5-shot\\
	\hline
	Full model &\textbf{0.262}&\textbf{0.317}&\textbf{0.401}&\textbf{0.452}\\
	\hline
	w/o distant neighbors &0.234&0.284&0.368&0.416\\
	w/o global encoder &0.203&0.207&0.270&0.361\\
	w/o local encoder &0.217&0.261&0.364&0.389\\
	\hline
\end{tabular}
\label{Ablat}
\end{table}

\begin{table*}[!tbp]
\caption{Evaluation results of 5-shot KGC on NELL-Only-LongTail and Wiki-Only-LongTail.}
\centering
\renewcommand\arraystretch{1.1}
\setlength{\tabcolsep}{2.2mm}{
	\begin{tabular}{c|cccc|cccc}
		\hline
		& \multicolumn{4}{c|}{NELL-Only-LongTail} & \multicolumn{4}{c}{Wiki-Only-LongTail} \\
		\cline{2-9}
		& MRR  &Hits@10  &Hits@5  &Hits@1 & MRR  &Hits@10  &Hits@5  &Hits@1\\
		\hline
		
		TransE &0.143&0.277&0.170&0.060 &0.118&0.125&0.092&0.055 \\
		TransH &0.151			
		&0.285&0.181&0.075  &0.113&0.125&0.084&0.042\\
		TransR &0.146			
		&0.275&0.164&0.080  &0.127&0.135&0.093&0.070\\
		RESCAL &0.251&0.266&0.235&0.135 &0.274&0.304&0.259&0.201 \\
		DisMult &0.223&0.241&0.216&0.170 &0.246&0.271&0.237&0.189 \\
		ComplEx &0.201&0.257&0.204&0.165 &0.263&0.299&0.251&0.191 \\ 
		\hline
		
		MetaR (In-Train) &0.321			
		&0.459 &0.405 &0.232  &0.240			
		&0.316 &0.285  &0.197\\
		MetaR (Pre-Train) &    0.260		
		&0.386 &0.336&	0.194 &0.344			
		&0.465 &0.423 &0.281\\
		FAAN &0.308		
		&0.439 &0.388 &0.233 & 0.328		
		&0.462 &0.403  &0.256\\
		\hline
		Full model &\textbf{0.347	}
		&	\textbf{0.466}  &\textbf{0.427}&	\textbf{0.281}  & \textbf{0.362}			
		&\textbf{0.487}&\textbf{0.429}& \textbf{0.295}\\	
		w/o distant neighbors &0.311	
		&0.435&	0.376 &	0.256  &0.328&0.434&0.403&0.261\\
		w/o global encoder &0.238		
		&0.341 &0.304 &0.178 &0.232&0.344&0.295&0.164\\					
		w/o local encoder &0.256		
		&0.427 &	0.353 & 0.176 &0.255 &0.423&0.367&0.186\\										
		
		\hline
\end{tabular}}
\label{long-tail-ablation}
\end{table*}
\begin{table*}[!tbp]
\caption{Results of RSCL and FAAN with 5-shot references on the test data of NELL-One. \#Avg.H and \#Avg.T denote the average numbers of direct neighbors of the head entities and the tail entities in the support sets, respectively.}
\centering
\renewcommand\arraystretch{1.1}
\setlength{\tabcolsep}{3.8mm}{
	\begin{tabular}{cccc|cc|cc}
		\hline
		&& & & \multicolumn{2}{c}{MRR} & \multicolumn{2}{c}{Hits@10} \\
		\cline{5-8}
		ID&Relations & \#Avg.H & \#Avg.T &RSCL &FAAN &RSCL &FAAN \\
		\hline
		1& AnimalSuchAsInvertebrate &1.0&2.2&\textbf{0.465}&0.391&\textbf{0.644}&0.605\\
		2 & AthleteInjuredHisBodypart &4.2&4.6&\textbf{0.613}&0.428&\textbf{0.734}&0.656\\
		3 & AutomobilemakerDealersInCity &2.0&2.8&0.062&\textbf{0.073}&0.110&\textbf{0.121}\\
		4&   ProducedBy &2.4&46.4&\textbf{0.570}&0.488&\textbf{0.812}&0.659\\
		5& PoliticianEndorsesPolitician &1.8&109.0&\textbf{0.248}&0.220&\textbf{0.339}&0.325\\
		6 &  AgriculturalProductFromCountry &2.4&119.2&\textbf{0.150}&0.139&\textbf{0.384}&0.296\\
		7  &  AutomobilemakerDealersInCountry &2.8&73.6&0.553&\textbf{0.597}&\textbf{0.868}&0.846\\
		8 & SportsGameSport &1.2&1263.0
		&\textbf{0.973}&0.965&\textbf{0.986}&0.977 \\
		9 &   TeamCoach &27.2&4.0&\textbf{0.096}&0.073&\textbf{0.167}&0.149\\  
		10 &GeopoliticalLocationOfPersion &10.0&7.2&\textbf{0.008}&\textbf{0.008}&\textbf{0.007}&\textbf{0.007}\\
		11 &  SportSchoolInCountry &25.0&278.6&0.420&\textbf{0.577}&0.684&\textbf{0.745}\\
		\hline
\end{tabular}}
\label{CaseStudy}
\end{table*}

\subsection{Results on long-tailed entities}
In this subsection, our goal is to validate the ability of our model to perform few-shot KG completion for long-tailed entities.
To this end, we construct two sub-datasets:
NELL-Only-LongTail and Wiki-Only-LongTail, which are composed of the triples that at least one of the head and tail entities is long-tailed\footnote{In this work, entities with less than $10$ direct neighbors are considered as long-tailed entities.}.
Based on the sub-datasets, 
we compare our full model with 
the three variant models described in Section \ref{sec-abla} (i.e., w/o distant neighbors, w/o global encoder and w/o local encoder), the traditional KG embedding baselines and the best performing FKGC baselines (i.e., MetaR and FAAN).
The results of 5-shot KG completion are summarized in Table \ref{long-tail-ablation}.
From the table, we have the following observations.
\begin{itemize}

\item Compared to the traditional KG baselines and the FKGC baselines, our full model achieves the best performance on both subdatasets, verifying the powerful ability of performing few-shot KG completion for long-tailed entities.

\item As we expected, our full
model significantly outperforms its three variant models on both sub-datasets, validating the necessity of encoding global and local neighborhood information when handling long-tailed entities. 
Note that our model without the global encoder achieves the worst performance among the three variants, further highlighting the importance of capturing graph context information.
\end{itemize}

\subsection{Case Study for Different Relations}

\begin{figure*}[!t]
\centering	
\subfigure{\includegraphics[width=0.95\linewidth, trim=0 0 0 0, clip]{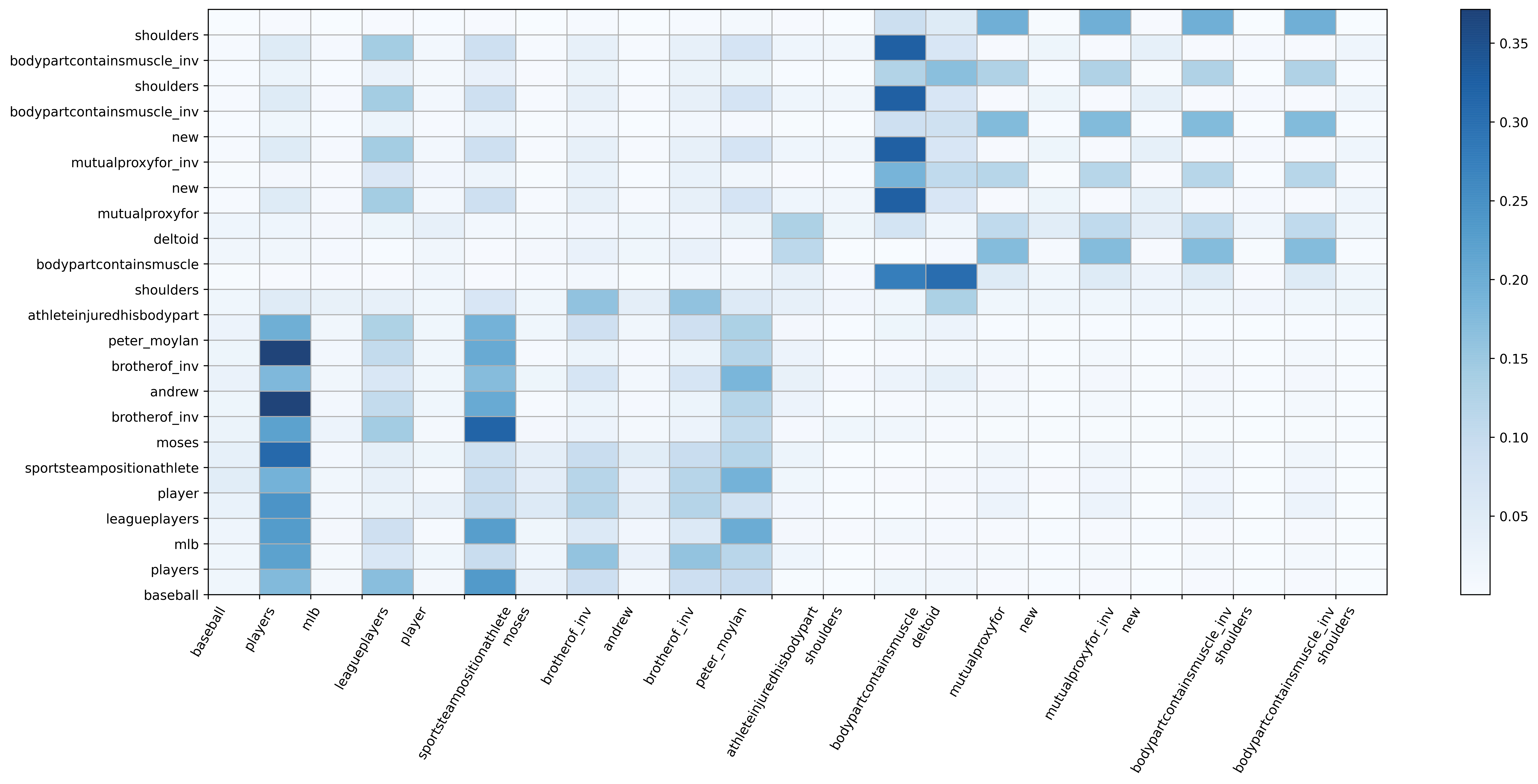}}
\caption{Visualization of attention map from self-attention in the global context encoder.
	The entity-relation sequence corresponds to the triple  (\texttt{peter\_moylan}, \texttt{AthleteInjuredHisBodypart}, \texttt{shoulders}).
	In the figure, \texttt{peter\_moylan} has five direct neighbors, and \texttt{shoulders} has one direct neighbor.
}
\label{attn_map}
\end{figure*}

To obtain a clearer view of the effectiveness of our method, we conduct a case study to analyze the results of different relations 
and the average numbers of direct neighbors of the head and tail entities in the reference triples.
Table \ref{CaseStudy} shows the decomposed results of RSCL and FAAN on NELL-One with 5-shot references.
We make the following observations.
\begin{itemize}
\item In the first ten rows of the table, the reference triples of the relations include at least one (head or tail) entity of which the average number of direct neighbors is no more than $10$.
This phenomenon demonstrates that
the reference triples used for training FKGC models often contain many long-tailed entities.
\item Compared with FAAN, RSCL generally improves the results of the few-shot relations of which reference triples contain long-tailed entities.
For example, for \texttt{AthleteInjuredHisBodypart}, RSCL obtains an MRR of $0.613$ and an Hits@10 of $0.734$,
which are $0.185$ and $0.078$ higher than FAAN, respectively.
The overall advance of RSCL indicates that our method is better at implementing FKGC for the few-shot relations having many long-tailed entities than FAAN.
We also notice that RSCL obtains lower results than FAAN on \texttt{SportSchoolInCountry}, which may be because the maximum value of neighbors set in the subgraph extractor limits the ability to handle the entities with a large number of direct neighbors.
\item The results of different relations are of high variance.
It can be seen that both RSCL and FAAN have good performance on some relations, such as \texttt{AutomobilemakerDealersInCountry} and \texttt{SportsGameSport}.
After analyzing their query and reference triples, we have two findings:
(1) Most of entities in the query triples have appeared in the reference triples, implying that existing FKGC models are more likely to predict the entities seen during training.
(2) The two relations usually have relatively small candidate entity sets (123 and 1,084 candidate entities), which makes their triples easier to predict.
In addition, for \texttt{GeopoliticalLocationOfPersion} that both RSCL and FAAN perform poorly on, we find that it has a very large number of candidate entities (up to 11,618), and the entities of its query triples hardly appear in the reference triples.
This suggests that the FKGC task is still far from being solved, and more intrinsic information about few-shot relations needs to be explored.


\end{itemize}

\subsection{Visualization}

\begin{figure}[!t]
\centering	
\subfigure{\includegraphics[width=8.5cm]{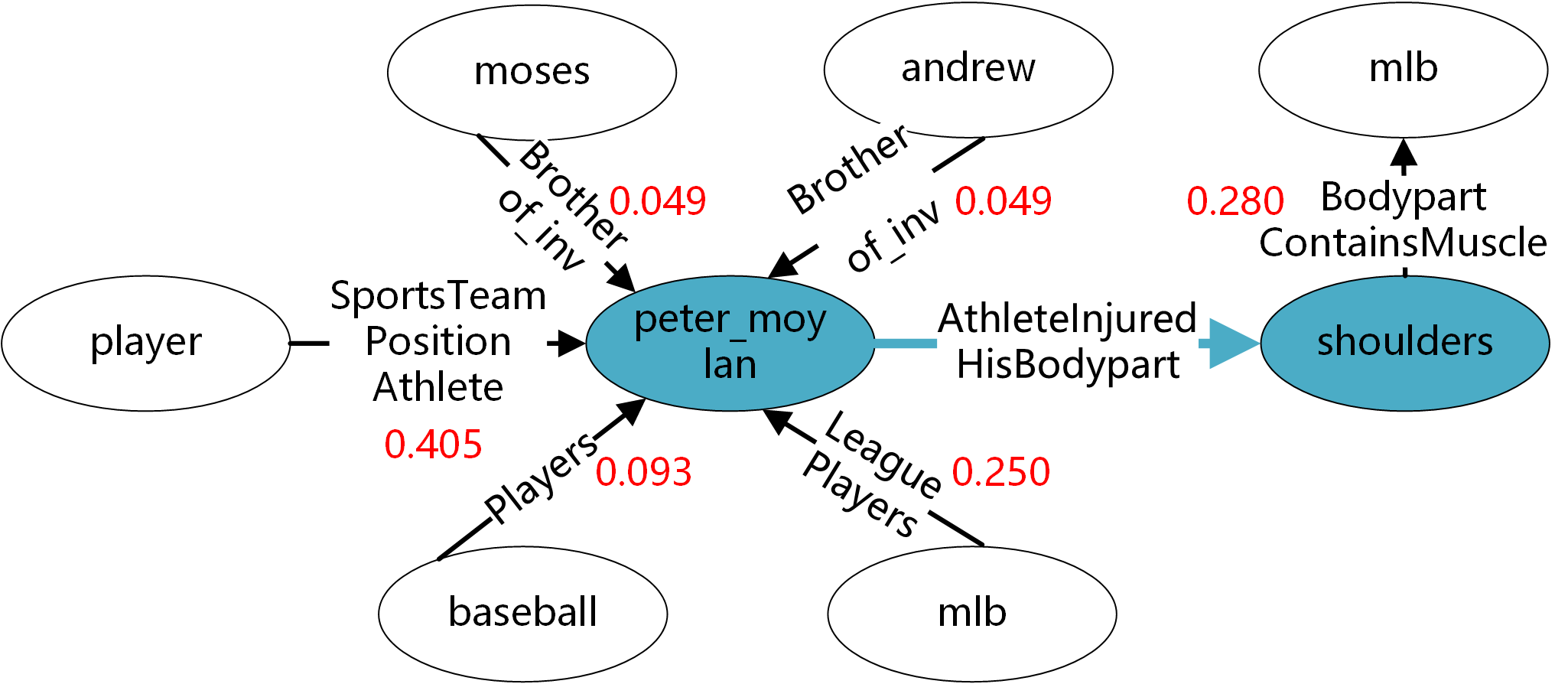}}
\caption{Visualization of the direct neighbors of (\texttt{peter\_moylan}, \texttt{AthleteInjuredHisBodypart}, \texttt{shoulders}) with their weights calculated by the local neighbor encoder.}
\label{local_attn}
\end{figure}

To intuitively understand the effect of the hierarchical attention network,
we visualize the attention weights of graph contexts.
Figure \ref{attn_map} displays the attention map of the global context encoder.
We observe that the few-shot relation \texttt{AthleteInjuredHisBodypart} has high attention scores connected to the distant directs of \texttt{shoulders}, which illustrates the effectiveness of the global context encoder for capturing long-term entity-relation dependencies.
Figure \ref{local_attn} depicts the direct neighbors of the triple with their weights calculated by the local neighbor encoder.
We observe that 
(\texttt{player}, \texttt{SportsTeamPositionAthlete}) has the largest weight for representing the triple (\texttt{peter\_moylan}, \texttt{AthleteInjuredHisBodypart}, \texttt{shoulders}), and (\texttt{andrew}, \texttt{Brotherof\_inv}) and (\texttt{moses}, \texttt{Brotherof\_inv}) are given the smallest weight. 
It is reasonable since \texttt{SportsTeamPositionAthlete} is obviously more relevant to \texttt{AthleteInjuredHisBodypart} than \texttt{Brotherof\_inv}.
Therefore, the local neighbor encoder can effectively capture local relation-specific information.

\section{Conclusion and future work}
In this paper, we proposed a Relation-Specific Context Learning (RSCL) method for FKGC.
Previous studies learned representations of few-shot relations based on the head and tail entities and the direct neighborhoods of entities.
RSCL focuses on learning global and local relation-specific representations for each few-shot relation by modeling graph contexts of triples, which can provide richer relation dependencies than direct neighborhoods without losing the valuable local information of entities.
The experimental results have revealed that RSCL outperforms existing state-of-the-art baselines on NELL-One and Wiki-One.
The ablation study and the case study for different relations illustrated the significant effect of the proposed hierarchical attention network.
In the future, we will consider both long-tailed entities and few-shot relations simultaneously and plan to explore external knowledge such as textual descriptions of entities and relations to solve the FKGC task.
Considering that Transformer is notoriously time-consuming, we will study low-cost lightweight networks that maintain comparable performance with less computational cost.


%
%
%
%
%
%
%
%
%
%
\ifCLASSOPTIONcaptionsoff
\newpage
\fi
\bibliographystyle{IEEEtran}
\bibliography{refs}
%

\end{document}